# A Standard Approach for Optimizing Belief Network Inference using Query DAGs

**Adnan Darwiche**
**Department of Mathematics**
**American University of Beirut**
**PO Box 11 - 236**
**Beirut, Lebanon**
*darwiche@aub.edu.lb*

**Gregory Provan**
**Department of Diagnostics**
**Rockwell Science Center**
**1049 Camino Dos Rios**
**Thousand Oaks, Ca 91360**
*provan@risc.rockwell.com*

## Abstract

This paper proposes a novel, algorithm-independent approach to optimizing belief network inference. Rather than designing optimizations on an algorithm by algorithm basis, we argue that one should use an *unoptimized* algorithm to generate a Q-DAG, a compiled graphical representation of the belief network, and then optimize the Q-DAG and its evaluator instead. We present a set of Q-DAG optimizations that supplant optimizations designed for traditional inference algorithms, including zero compression, network pruning and caching. We show that our Q-DAG optimizations require time linear in the Q-DAG size, and significantly simplify the process of designing algorithms for optimizing belief network inference.

## 1 Introduction

Query DAGs (Q-DAGs) have been introduced recently to allow the cost-effective implementation of belief network inference on multiple software and hardware platforms [1, 2]. According to the Q-DAG approach, belief network inference is decomposed into two steps as shown in Figure 1. The first step takes place off-line and results in the generation of a Q-DAG that can answer a number of pre-specified probabilistic queries. The second step takes place on-line and involves the evaluation of a Q-DAG to compute answers to probabilistic queries in the context of some given evidence. A Q-DAG evaluator is a very simple piece of software, which allows one to implement it cost-effectively on multiple software and hardware platforms.

Our initial discussion of Q-DAGs has focused on three key points: (a) Q-DAGs can be generated using modified versions of standard belief network algorithms; (b) the time and space complexity of Q-DAG generation is the same as the time complexity of the underlying belief network algorithm; and (c) a Q-DAG evaluator is a very simple piece of software [1, 2].

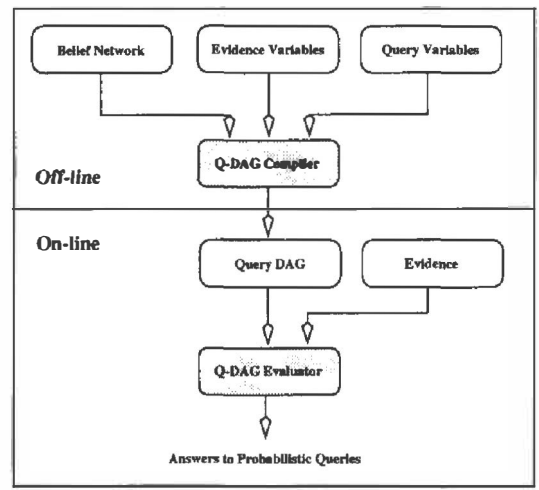

Figure 1: The Query DAG framework.

Our own experience, however, has revealed another important property of Q-DAGs that was not originally intended but that seems to be as crucial as the multiple-platform feature. In a nutshell, when using a belief network algorithm to generate a Q-DAG, one need not worry about optimizing the algorithm using techniques such as computation-caching, zero-compression, and network-pruning [3, 4, 5]. Similar, if not better, efficiency can be expected by simply using an unoptimized version of the algorithm to generate a Q-DAG and then optimizing inference at the Q-DAG level. This involves reducing the Q-DAG before evaluating it and implementing an optimized Q-DAG evaluator. The same Q-DAG evaluator can be used with any generation algorithm and optimizations at the Q-DAG level seem to be much simpler to understand and implement since they deal with graphically represented arithmetic expressions, without having to invoke prob-ability or belief network theory. Therefore, the merits of this alternative approach are many, but most importantly: the Q-DAG alternative is systematic, simple and accessible to a bigger class of algorithms and developers.

The rest of this paper is structured as follows. First, we will review Q-DAGs and their semantics in Section 2. Next, we will present complete pseudocode for an optimized Q-DAG evaluator in Section 3 and discuss its computational complexity. In Section 4, we present techniques and pseudocode for reducing the size of a Q-DAG and discuss its computational complexity. Section 5 is then dedicated to how Q-DAG reduction and evaluation account for many of the standard optimization techniques that one seeks to realize in belief network algorithms. Moreover, we will argue in this section that the Q-DAG approach is not just an alternative, but a better alternative according to a number of measures that we shall also discuss. We finally close in Section 6 with some concluding remarks.

## 2   Query DAGs

We will review Q-DAGs using an example. Consider the belief network in Figure 2(a) and suppose we are interested in queries of the form $Pr(b \mid c)$. Figure 2(c) depicts a Q-DAG for answering such queries, which is essentially a parameterized arithmetic expression where the value of parameters depend on the evidence obtained. This Q-DAG will actually answer queries of the form $Pr(b, c)$, but we can use normalization to compute $Pr(b \mid c)$. This Q-DAG was generated using the join tree algorithm which builds a join tree as shown in Figure 2(b). Details of this generation process can be found in [1, 2].

A number of observations are in order this Q-DAG:

- It has two leaf nodes labeled $(B, ON)$ and $(B, OFF)$. These are called *query nodes* because their values represent answers to queries $Pr(B=ON, e)$ and $Pr(B=OFF, e)$ where e is the available evidence.

- It has two root nodes labeled $(C, ON)$ and $(C, OFF)$. These are called *Evidence-Specific Nodes (ESNs)* since their values depend on the evidence collected about variable $C$ on-line.

According to the semantics of Q-DAGs, the value of evidence–specific node $(V, v)$ is 1 if variable $V$ is *observed* to have value $v$ or is *unknown,* and 0 otherwise. Once the values of ESNs are determined, we evaluate the remaining nodes of a Q-DAG using numeric multiplication and addition. The numbers that get assigned to query nodes as a result of this evaluation are the answers to queries represented by these nodes.

For example, suppose that the evidence we have is $C=ON$. Then the value of ESN $(C, ON)$ is set to 1 and the value of ESN $(C, OFF)$ is set to 0. The Q-DAG in Figure 2(c) is then evaluated as given in Figure 3(a), thus leading to $Pr(B=ON, C=ON) = .3475$ and $Pr(B=OFF, C=ON) = .2725$,. If the evidence we have is $C=OFF$, however, then $(C, ON)$ is set to 0 and $(C, OFF)$ is set to 1. The Q-DAG in Figure 2(c) will then be evaluated as given in Figure 3(b), thus leading to $Pr(B=ON, C=OFF) = .2875$ and $Pr(B=OFF, C=OFF) = .0925$.

If we have no evidence about variable $C$ (the value of $C$ is unknown), both evidence–specific nodes $(C, ON)$ and $(C, OFF)$ will then be set to 1 and the remaining nodes will be evaluated accordingly.

Formally, a probabilistic Q-DAG is a directed acyclic graph. Each root node in a Q-DAG is either a *numeric node, Num,* which is labeled with a number $p$ in $[0, 1]$, or an *evidence–specific node, Esn,* which is labeled with a pair $(V, v)$ where $V$ is a variable and $v$ is a value of the variable. Each non–root node is either a *multiplication node,* $\otimes$, which is labeled with a $*$, or an *addition node,* $\oplus$, which is labeled with a $+$.

In the rest of this paper, we assume the following functions for manipulating Q-DAGs: *Children(n):* the children of node $n$; *Parents(n):* the parents of node $n$; *Type(n):* the type of node $n$, which is either $\otimes$, $\oplus$, *Esn* or *Num; Label(n):* the probability associated with a numeric node $n$; *Value(n):* the value of a Q-DAG node $n$. Given *Values* for evidence–specific nodes, the *Value* of nodes can be determined as follows. The *Value* of a numeric nodes is its *Label;* the *Value* of an addition node is the addition of its parents' *Values;* the *Value* of a multiplication node is the multiplication of its parents' *Values.*

We will also use $\mathcal{N}$ to represent the number of nodes in a Q-DAG and $\mathcal{E}$ to represent the number of edges.

## 3   A Q-DAG Evaluator

We now discuss an optimized Q-DAG evaluator that initializes the probabilities (values) of Q-DAG nodes and updates them as evidence changes. Evidence in this case is the setting of an evidence–specific node to either 0 or 1 according to Q-DAG semantics described in [1, 2] and reviewed in Section 2.

There are two high level procedures for implementing the evaluator, the goal of which is to keep the function *Value* updated. This function assigns a probability *Value(n)* to each node $n$ so that if $n$ is a query node

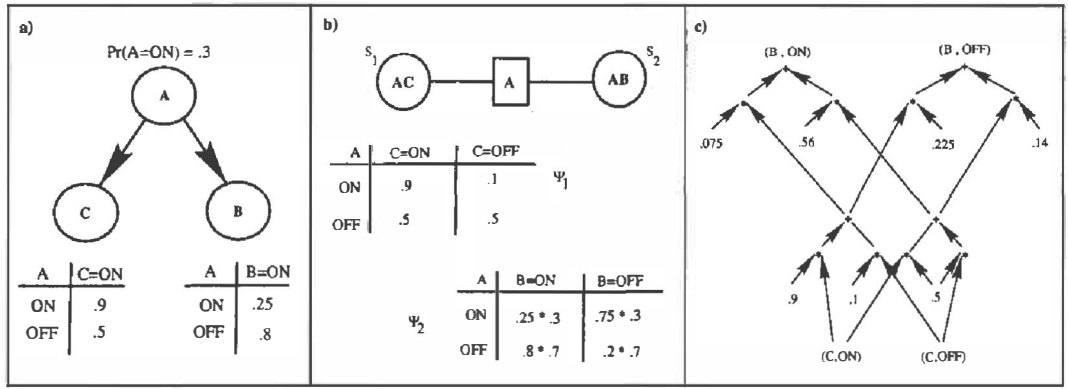

Figure 2: (a) A belief network; (b) a corresponding join tree; and (c) a Q-DAG generated using the join tree.

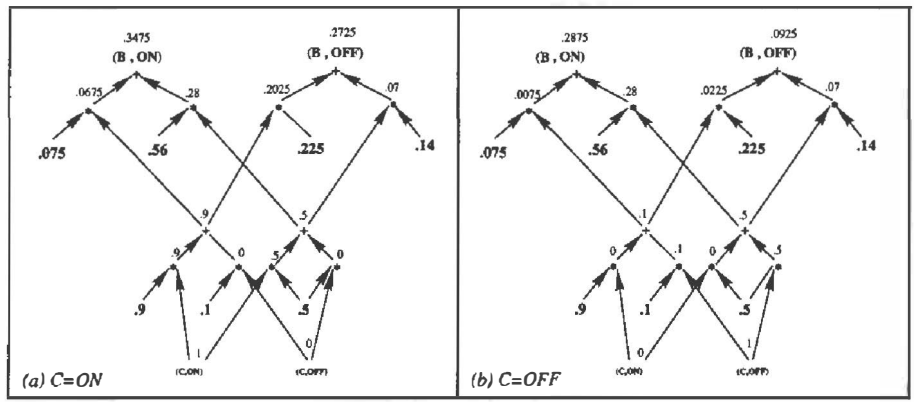

Figure 3: Q-DAG evaluation.

labeled with $(V, v)$, and if $Value(n) = p$, then we must have $Pr(v, e) = p$. The two procedures are:

1. **initialize-qdag** (Figure 4): computes probabilities of nodes under the assumption that no evidence is collected (all evidence–specific nodes are set to 1).

2. **set-evidence** (Figure 5): sets the value of an evidence–specific node $n$ to $v$ and updates the function $Value$ accordingly.

One would use these procedures by first calling **initialize-qdag** to initialize the Q-DAG. As evidence becomes available, corresponding calls to **set-evidence** are made.

Procedure **initialize-qdag** starts by initializing the probability of each query node. To initialize the probability of a node $n$, one first initializes the probabilities of its parents recursively and then combines these probabilities depending on the type of node $n$. The boundary conditions occur when $n$ is a root node (*Esn* or *Num*). If $n$ is an evidence–specific node, its initial value is 1.[1] And if $n$ is a numeric node, then its initial value is simply the label associated with it. The initialization algorithm takes $O(\mathcal{E})$ time and it computes the prior probability of each query node.[2]

Procedure **set-evidence** works by incrementally updating the probabilistic values of nodes. Specifically, suppose that the value of node $n$ changes from $v_1$ to $v_2$ and consider a child $m$ of $n$:

- If $m$ is an addition node, then its value will change by the amount $v_2 - v_1$, which is also the change that node $n$ has undergone. Therefore, we can update the value of node $m$ by simply adding $v_2 - v_1$ to its previous value. We must also update the children of $m$ recursively.

- If $m$ is a multiplication node, and if $v_1 \neq 0$, we can

[1]When no evidence is available, all evidence–specific nodes have the value 1.

[2]Note that initialization can be done off-line since it does not depend on any evidence. Therefore, it should not, in principle, be part of the Q-DAG evaluator but part of the Q-DAG compiler.

```
initialize-qdag()
    for every query node n do initialize-prob(n)

initialize-prob(n)
    unless Value(n) is initialized do
        for every node m in Parents(n) do
            initialize-prob(m)
        case Type(n)
            Num : Value(n) := Label(n)
            Esn : Value(n) := 1
            ⊗ : Value(n) := value-of-mul-node(n)
            ⊕ : Value(n) := value-of-add-node(n)

value-of-mul-node(n)
    Value := 1
    for all m in Parents(n) do
        Value := Value * Value(m)
    return Value

value-of-addition-node(n)
    Value := 0
    for all m in Parents(n) do
        Value := Value + Value(m)
    return Value
```

Figure 4: Initializing a Q-DAG.

update the value of node $m$ by simply multiplying $v_2/v_1$ by its previous value. If $v_1 = 0$, however, we cannot incrementally update the value of $m$. Instead, we have to recompute its value by multiplying the values of its parent nodes, which is what function value-of-mul-node does.

Procedure set-evidence takes $O(\mathcal{E})$ time in the worst case, but its average performance is better than linear since it will only visit those nodes that change their values.

## 4   Simplifying the Q-DAG

One may reduce a Q-DAG by eliminating some of its nodes and arcs while maintaining its ability to answer probabilistic queries correctly. The motivation behind this simplification or reduction is twofold: faster evaluation of Q-DAGs and less space to store them. Interestingly enough, we have observed that a few, simple reduction techniques tend in certain cases to subsume optimization techniques that have been influential in practical implementations of belief network systems. Therefore, reducing Q-DAGs can be very important practically.

This section is structured as follows. First, we start by discussing three simple reduction techniques in the form of rewrite rules. Next, we provide pseudocode that implements these reductions and discuss their computational complexity. Finally, the implications of these reductions on optimizing belief network infer-

```
set-evidence(n, NewValue)
    OldValue := Value(n)
    Value(n) := NewValue
    propagate-change(n, OldValue, NewValue)

propagate-change(n, OldValue, NewValue)
    unless OldValue = NewValue do
        for each node m in Children(n) do
            OldChildValue := Value(m)
            if Type(m) = ⊕
            then Value(m) :=
                    Value(m) − OldValue + NewValue
            else if OldValue = 0
                then Value(m) :=
                        value-of-mul-node(m)
                else Value(m) :=
                        Value(m)/OldValue * NewValue
            propagate-change(m, OldChildValue, Value(m))
```

Figure 5: An optimized Q-DAG evaluator.

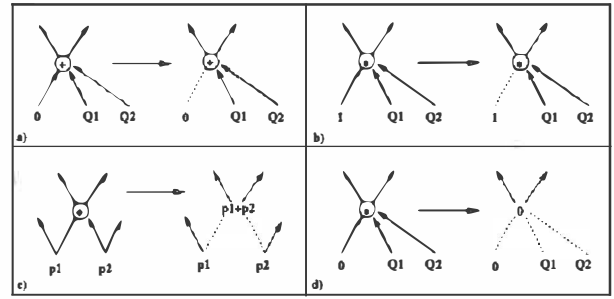

Figure 6: Q-DAG reduction techniques.

ence are discussed at length in Section 5.

The goal of Q-DAG reduction is to reduce the size of a Q-DAG while maintaining the arithmetic expression it represents.

**Definition 1** *Two Q-DAGs are equivalent iff they have the same set of evidence–specific nodes, the same set of query nodes, and they agree on the values of query nodes for all possible values of evidence–specific nodes.*

Figure 6 shows three basic reduction operations on Q-DAGs. *Identity–elimination* eliminates a numeric node if it is an identity element of its child (Figures 6(a) and 6(b)). *Numeric–reduction* replaces a multiplication or addition node with a numeric node if all its parents are numeric nodes (Figure 6(c)). *Zero–compression* replaces a multiplication node by a numeric node if one of its parents is a numeric node with value zero (Figure 6(d)).

Identity elimination is implemented by the straightforward procedures eliminate-identity-zero and eliminate-identity-one in Figure 7, each of which takes $O(\mathcal{N})$ time.

```
eliminate-identity-zero()
    for each node n do
        if Type(n) = Num and Label(n) = 0
        then for every node m in Children(n) do
                if Type(m) = ⊕
                then Parents(m) := Parents(m) \ {n}

eliminate-identity-one()
    for each node n do
        if Type(n) = Num and Label(n) = 1
        then for every node m in Children(n) do
                if Type(m) = ⊗
                then Parents(m) := Parents(m) \ {n}
```

Figure 7: Pseudocode for identity elimination.

```
numeric-reduction()
    initialize queue q
    for every node n do
        case Type(n)
            Num : add n to queue q
            ⊗ : InDegree(n) := | Parents(n) |
            ⊕ : InDegree(n) := | Parents(n) |
    while queue q is not empty do
        get n from queue q
        for each m in Children(n) do
            InDegree(m) := InDegree(m) - 1
            if InDegree(m) = 0
            then Type(m) := Num
                 Label(m) := Value(m)
                 Parents(m) := ∅
                 add m to queue q
```

Figure 8: Pseudocode for numeric reduction.

Numeric reduction is implemented by procedure **numeric-reduction** in Figure 8, which maintains a queue of numeric nodes in the Q-DAG and a counter for each addition/multiplication node to count its parents. For each (numeric) node on the queue, the procedure processes the node by decrementing the counters of its children. If any of these counters reaches zero, that means all parents of the corresponding nodes are numeric and, therefore, it can be reduced into a numeric node. When the reduction is performed, the node is added to the queue, which allows the possible reduction of its children. Procedure **numeric-reduction** takes $O(\mathcal{E})$ time.

Zero-compression is implemented by the procedure **zero-compression** in Figure 9. Note here that if any Q-DAG node attains the value zero after calling procedure **initialize-qdag**, it will maintain this value under any further evidence.[3] Procedure **zero-compression** is complete with respect to zero-

----

[3] Accommodating evidence entails changing the values of some evidence-specific nodes from 1 to 0. This cannot increase the value of any Q-DAG node.

```
zero-compression()
    for each node n do
        if Type(n) = Num
        then Type(n) := Num
             Value(n) = 0
             Label(n) := 0
             Parents(n) := ∅
```

Figure 9: Pseudocode for zero compression.

compression as depicted in Figure 6(d) because every multiplication node that has a zero parent will also have the value zero, and, therefore, will be converted to a numeric node. The time complexity of **zero-compression** is $O(\mathcal{N})$.

## 5    Optimization using Q-DAGs

The main proposal in this paper is as follows: Instead of implementing an optimized algorithm for belief network inference, use an unoptimized version of the algorithm to generate a Q-DAG, reduce the Q-DAG using the procedures in Figures 7–9, and evaluate it using the procedures in Figure 5.

The benefits of the Q-DAG approach are: (1) the Q-DAG evaluator can be easily and cost–effectively implemented on various software and hardware platforms; (2) Q-DAG reduction and evaluation are algorithm–independent; (3) Q-DAG reduction subsumes the technique of zero–compression, and some forms of network pruning; (4) the Q-DAG evaluator implements a sophisticated scheme for computation caching, which is simpler and more refined than any of the caching schemes that are typically implemented in algorithms based on message passing; (5) the Q-DAG evaluator handles the retraction of evidence with minimal computations, while most caching mechanisms we are aware of seem to have difficulties in handling this kind of evidence efficiently.

The first two points above are self evident and will not be discussed further. We focus in the rest of this section on the last three points, explaining them in detail and supporting them by examples.

### 5.1    Zero Compression

Zero compression is an optimization technique that is typically implemented in algorithms based on join trees [5]. Zero compression is designed to take advantage of conditional probability tables which contain zero entries, implying some logical or functional relationship between network variables. During initialization of a join tree, each zero conditional probability is multiplied into some clique entry, which causes the corresponding entry in the clique to be zero as well.

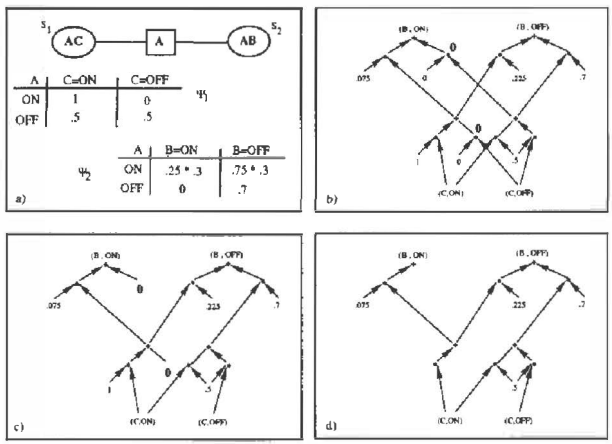

Figure 10: Zero compression at the Q-DAG level.

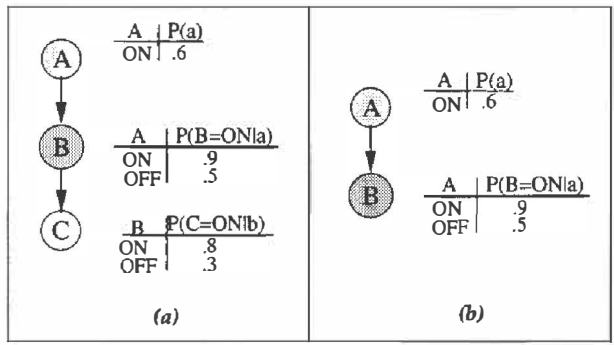

Figure 11: Pruning Node $C$.

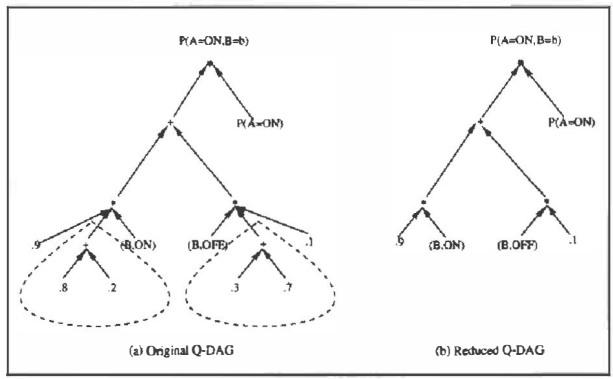

Figure 12: Reducing a Q-DAG.

After performing some message passing to propagate evidence, some of these zeros will propagate throughout the entire join tree. As more evidence is obtained and propagated, computational resources are expended adding and multiplying clique entries by zeros. Zero compression, as presented in [5], addresses this wasteful propagation by visiting entries in cliques to identify and annihilate the zero entries. The annihilation step should restructure the internals of cliques to exclude zero entries from subsequent message passes. The same optimization can also be implemented in the context of other algorithms, but the details would differ.

What is common, however, among different algorithms is that one can save computationally, sometimes significantly, by avoiding multiplications by zeros whenever possible. As we demonstrate now, this zero compression optimization is subsumed by our Q-DAG zero-compression technique that we discussed in Section 4.

Consider the Q-DAG in Figure 2. Suppose that $Pr(C = OFF \mid A = ON) = 0$ and $Pr(B = ON \mid A = OFF) = 0$ instead. The resulting join tree will then be as given in Figure 10(a) where each clique has a zero entry. The technique of zero compression aims at factoring out these entries so they do not enter into further computations when propagating messages.

Alternatively, one could use a join tree algorithm that does not incorporate zero compression to generate the Q-DAG shown in Figure 10(b). One would then initialize the Q-DAG to discover that some nodes will attain the value zero. Procedure `zero-compression` can then be applied to generate the Q-DAG in Figure 10(c), which could further be reduced using Procedure `eliminate-identity-zero` leading to the Q-DAG in Figure 10(d). Therefore, one need not worry about implementing zero compression in the chosen belief network algorithm; one can rely on Q-DAG reduction to achieve the same result as illustrated above.

## 5.2 Network Pruning

Pruning is the process of deleting irrelevant parts of a belief network before invoking inference. Consider the network in Figure 11(a) for an example, where $B$ is an evidence variable and $A$ is a query variable. One can prune node $C$ from the network, leading to the network in Figure 11(b). Any query of the form $Pr(a \mid b)$ will have the same value with respect to either network, but working with the smaller network is clearly preferred.

Now, if we generate a Q-DAG for the network in Figure 11(a) using the polytree algorithm, we obtain the one in Figure 12(a). On the other hand, if we generate a Q-DAG for the network in Figure 11(b), we obtain the one in Figure 12(b), which is smaller as expected. The key observation, however, is that the optimized Q-DAG in Figure 12(b) can be obtained from the unoptimized one in Figure 12(a) using Q-DAG reduction. In particular, the nodes enclosed in dotted lines can be collapsed using numeric–reduction into a single node with value 1. Identity–elimination can then remove the resulting node, leading to the optimized Q-DAG in Figure 12(b).

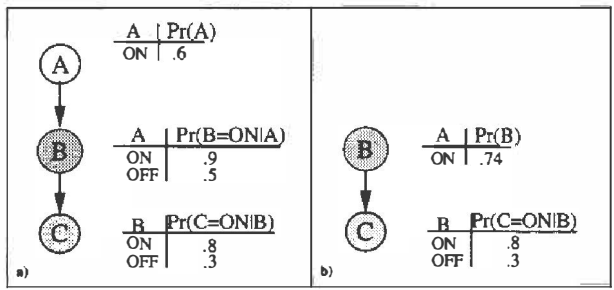

Figure 13: Pruning Node $A$.

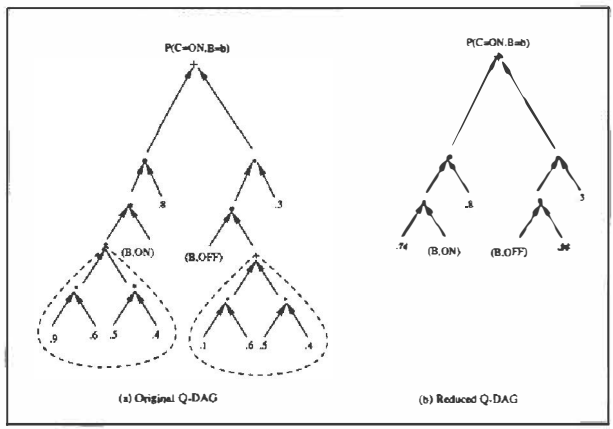

Figure 14: Reducing a Q-DAG.

To consider another example, suppose that we are interested in computing $Pr(C = ON \mid b)$ in the network of Figure 13(a). One can prune node $A$ from the network, leading to the network in Figure 13(b) where the priors of node $B$ are computed as follows: $Pr(b) = \sum_a Pr(b \mid a)Pr(a)$. Any query of the form $Pr(c \mid b)$ will have the same value with respect to either network, but working with the smaller network is clearly referred.

If we generate a Q-DAG for the network in Figure 13(a) using the polytree algorithm, we obtain the one in Figure 14(a). If we generate a Q-DAG for the network in Figure 13(b), we obtain the one in Figure 14(b). Note, however, that the Q-DAG in Figure 14(b) can be obtained from the Q-DAG in Figure 14(a) using numeric–reduction.

Therefore, some forms of network pruning are a by-product of Q-DAG reduction and, hence, one can decide to ignore them at the algorithmic level and expect that their effect will be realized if Q-DAG reduction is utilized. There are two caveats, however. First, it is not clear whether all forms of network pruning will be subsumed by Q-DAG reduction. Second, Q-DAG reduction will not reduce the computational complexity of inference, although network pruning may. For example, a multiply–connected network may become

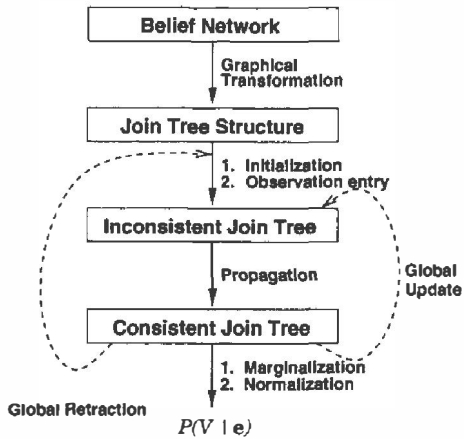

Figure 15: Block diagram of the join tree algorithm.

singly–connected after pruning, thereby, reducing the complexity of inference. But using Q-DAG reduction, we still have to generate a Q-DAG using the multiply-connected network.

## 5.3  Dynamic Evidence

The proper handling of dynamic evidence is an essential property of practical belief network inference. Inference with dynamic evidence is typically implemented with a computation caching scheme that attempts to maximize the reuse of previous computations to conduct new ones. Unfortunately, computation caching is non-trivial, typically undocumented, and its details vary from one algorithm to another.

The main objective of this section is to show that a Q-DAG framework allows us to handle dynamic evidence in a simple, uniform and sophisticated manner. Using this framework, we can (a) ignore dynamic evidence at the algorithmic level, (b) use the algorithm to generate a Q-DAG, and (c) handle dynamic evidence at the Q-DAG level. But before we substantiate these claims, we review how dynamic evidence is typically handled in the join tree algorithm [3, 4, 5].

Figure 15, which is borrowed from [3], depicts the overall control of the join tree algorithm. There are two important points to notice about this figure. First, the introduction of evidence leads to invalidating certain computations, which leads to an *inconsistent join tree*. The goal is then to recover this consistency (validate probabilities) by doing the least amount of work possible. Second, there is a distinction between *evidence update* and *evidence retraction*, in that evidence retraction requires more work to accommodate.

We apply *evidence update* to variable $V$ if its current value is unknown but evidence suggests a new value $v$ of $V$. We apply *evidence retraction* to variable $V$ if it has a current value $v_1$ but evidence retracts this value or suggests a different value $v_2$. In the join tree algorithm, evidence update requires recomputing certain messages which are passed between cliques. Moreover, the messages to be recomputed are decided upon by *certain flags that indicate the validity* of messages as evidence is collected. Evidence retraction requires *in addition* the re-initialization of certain clique potentials. Details of these operations are beyond the scope of this paper, but see [3] for a relatively comprehensive discussion. The metric we use for determining how well a system handles dynamic evidence is the amount of work needed to update probabilities.

First, we need to define evidence update and retraction formally in the context of a Q-DAG framework:

**Definition 2** *Evidence update* *occurs when each evidence–specific node either maintains its value or changes its value from 1 to 0. Evidence retraction occurs when some evidence–specific node changes its value from 0 to 1.*

Given the Q-DAG semantics of Section 2, evidence update occurs if and only if each variable either maintains its observed value or changes from unknown to observed. On the other hand, evidence retraction occurs if and only if some observed variable either becomes unknown or changes its observed value to a different one.

Dynamic evidence is handled in the Q-DAG framework as follows. Both evidence update and retraction are handled using the same procedure **set-evidence** given in Figure 5. As far as *simplicity* is concerned, the pseudocode in Figure 5 speaks for itself. As far as *uniformity* is concerned, this code is independent of the algorithm used to generate the Q-DAG; therefore, it can be used with any Q-DAG generation algorithm. As far as *efficiency* is concerned, we have three points to make. First, **set-evidence** takes $O(\mathcal{E})$ time in the worst case but does much better on average since it only visits nodes that change their values. Second, the caching scheme implied by **set-evidence** is more refined than schemes based on message passing. Note that each message pass involves a number of arithmetic operations which correspond to some Q-DAG nodes. If a message becomes invalid, all of these operations must be re-applied although some of them may not lead to new values. In a Q-DAG framework, only nodes that change their values are re-evaluated, therefore leading to a more refined caching scheme. This level of refinement is missed in message passing algorithms since caching is done at the message level, not at the arithmetic operation level. Our final point is regarding the minor difference between evidence update and retraction in the Q-DAG framework, which is contrary to what one finds in other frameworks. Specifically, in evidence update, the condition $OldValue = 0$ will never be satisfied when calling the procedure **set-evidence** and procedure **value-of-mul-node** will never be invoked. This follows since no node will increase its value given that no evidence–specific node has increased its value (in evidence update, the value of an evidence–specific node will never change from 0 to 1). The procedure **value-of-mul-node** may only be invoked in case of evidence retraction, which is the only extra work needed to handle evidence retraction versus evidence update.

## 6    Conclusion

The message of this paper is simple: instead of optimizing belief network algorithms, (a) use plain, unoptimized versions of the algorithms to generate a Q-DAG, (b) reduce the Q-DAG according to the procedures given in Figures 7–9; and (c) evaluate the Q-DAG using the procedures given in Figure 5. This proposed alternative is cost–effective, uniform, relatively simple, and optimized as compared to the standard approach.